\documentclass{article}
\usepackage{iclr2027_conference,times}

\usepackage[T1]{fontenc}
\usepackage{hyperref}
\definecolor{citecolor}{HTML}{0071bc}

\hypersetup{colorlinks=true, linkcolor=citecolor, citecolor=citecolor, urlcolor=citecolor}
\usepackage{url}
\usepackage{graphicx}
\usepackage{wrapfig}
\usepackage{xspace}
\usepackage{amsmath}
\usepackage{amssymb}
\usepackage{amsthm}

\usepackage{booktabs}
\usepackage{multirow}
\usepackage{subcaption}
\usepackage{xcolor}
\usepackage{colortbl}
\usepackage{array}
\usepackage{tikz}
\usetikzlibrary{positioning,arrows.meta,calc,fit,backgrounds}
\usepackage{algorithm}
\usepackage[noend]{algpseudocode}
\usepackage{listings}
\usepackage[most]{tcolorbox}

\graphicspath{{images/}}

\newcommand{\modelTeacher}{\textit{DeepSeek V4 Pro}\xspace}
\newcommand{\modelStudent}{\textit{Qwen~3.5~9B}\xspace}
\newcommand{\modelStudentLarge}{\textit{Qwen~3.5~27B}\xspace}
\newcommand{\modelSol}{\textit{Sol Pro}\xspace}
\newcommand{\modelSolFull}{\textit{GPT-5.6} \modelSol}
\newcommand{\modelLuna}{\textit{Luna}\xspace}
\newcommand{\modelLunaFull}{\textit{GPT-5.6} \modelLuna}
\newcommand{\modelGrok}{\textit{Grok-4.6}\xspace}
\newcommand{\modelMiniMax}{\textit{MiniMax-m3}\xspace}

\newcommand{\datasetDSVPro}{DSv4Pro\xspace}
\newcommand{\datasetNemotron}{Nemotron~v3\xspace}

\newcommand{\benchmarkSWEVerified}{SWE-bench Verified\xspace}
\newcommand{\harnessMiniSWE}{mini-swe-agent\xspace}

\newcommand{\metacurate}{\textsc{MetaCurate}\xspace}

\newcommand{\baselineBase}{Base\xspace}
\newcommand{\baselineFullPool}{Full pool\xspace}
\newcommand{\baselineRandom}{Random\xspace}
\newcommand{\baselineResolved}{Resolved\xspace}
\newcommand{\baselineGoldPatchRecall}{Gold-patch recall\xspace}
\newcommand{\baselineRuleFilter}{Rule filter\xspace}
\newcommand{\baselineLLMFilter}{LLM filter\xspace}
\newcommand{\baselineCurateEvo}{CurateEvo\xspace}
\newcommand{\methodOurs}{Ours\xspace}

\newcommand{\termmark}[1]{#1}
\newcommand{\termword}[1]{\termmark{#1}\xspace}

\newcommand{\instrFs}{\termword{curation filters}}
\newcommand{\instr}{\termword{filter}}
\newcommand{\instrs}{\termword{filters}}

\newcommand{\InstrT}{Curation Filter\xspace}
\newcommand{\InstrsT}{Curation Filters\xspace}

\definecolor{takeawayframe}{HTML}{3F5166}
\definecolor{takeawayback}{HTML}{F2F6FB}
\newtcolorbox{takeawaybox}{
  enhanced, breakable, sharp corners=downhill, arc=1.5pt,
  colback=takeawayback, colframe=takeawayframe, boxrule=0.5pt,
  left=6pt, right=6pt, top=4pt, bottom=4pt,
  before skip=6pt, after skip=4pt}
\newcommand{\takeaway}[1]{\begin{takeawaybox}\textbf{Design implications.}\ #1\end{takeawaybox}}

\definecolor{ourshade}{HTML}{E4EEF9}
\newcommand{\oursrow}{\rowcolor{ourshade}}

\definecolor{promptframe}{HTML}{3F5166}
\definecolor{promptback}{HTML}{F7F8FA}
\definecolor{promptarrow}{HTML}{8A97A6}
\lstdefinestyle{promptstyle}{
  basicstyle=\ttfamily\scriptsize,
  breaklines=true,
  breakindent=1em,
  postbreak=\mbox{\textcolor{promptarrow}{$\hookrightarrow$}\space},
  columns=fullflexible,
  keepspaces=true,
  showstringspaces=false,
  upquote=true,
  aboveskip=0pt, belowskip=0pt,
  xleftmargin=0pt, framexleftmargin=0pt,
}
\newtcolorbox{promptbox}[1]{
  breakable, enhanced, title={#1},
  colback=promptback, colframe=promptframe,
  fonttitle=\bfseries\footnotesize, boxrule=0.6pt, arc=2pt,
  left=4pt, right=4pt, top=3pt, bottom=3pt,
  toptitle=1.5pt, bottomtitle=1.5pt}

\title{\metacurate: Evolving \InstrsT for High-Quality Agentic Coding Data
}

\author{
\begin{minipage}{\dimexpr\textwidth-3\tabcolsep\relax}
\raggedright\normalfont
\textbf{Donghoon~Kim}\textsuperscript{1\raisebox{0.35ex}{*}}, \textbf{Seunghyuk~Oh}\textsuperscript{1\raisebox{0.35ex}{*}},
\textbf{Kevin~Galim}$^{1}$, \textbf{Minseo~Kim}$^{2}$,
\textbf{Minjae~Lee}$^{1}$, \textbf{Younghyun~Kim}$^{1}$,
\textbf{Wooseok~Seo}$^{1}$, \textbf{Hyung~Il~Koo}$^{1}$,
\textbf{Wonjun~Kang}$^{1}$
\endgraf\medskip
{\small $^{1}$FuriosaAI \quad $^{2}$University of California, Berkeley}
\endgraf\vspace{0.2em}
{\small Project page: \href{https://furiosa-air.github.io/metacurate}{\texttt{\footnotesize furiosa-air.github.io/metacurate}}\hspace{0.8em}
\\Code: \href{https://github.com/furiosa-ai/MetaCurate}{\texttt{\footnotesize github.com/furiosa-ai/MetaCurate}}\hspace{0.8em}
\\Dataset: \href{https://huggingface.co/datasets/furiosa-ai/MetaCurate-DeepSeekV4Pro}{\texttt{\footnotesize huggingface.co/datasets/furiosa-ai/MetaCurate-DeepSeekV4Pro}}}
\end{minipage}
}
\iclrfinalcopy

\begin{document}

\maketitle
\lhead{Preprint}
\begingroup
\renewcommand{\thefootnote}{\fnsymbol{footnote}}
\footnotetext[1]{Equal Contribution. Emails: \texttt{\{dhkim, seunghyukoh\}@furiosa.ai}.}
\endgroup

\begin{abstract}

Supervised fine-tuning (SFT) on agent trajectories is an effective way to improve coding agents, but selecting useful training examples requires more than identifying successful runs.
A resolved patch may follow inefficient reasoning or unnecessary tool use, while a failed run may still contain useful intermediate steps.
Training pipelines therefore rely on rules and qualitative assessments to filter trajectories, yet designing these filters typically requires substantial human expertise, and validating improvements through repeated training and evaluation is expensive.
We introduce \metacurate, a training-free framework that evolves executable trajectory curation filters using pairwise LLM judgments.
Disagreements between the filter's rankings and the judge's preferences provide targeted feedback for jointly revising its rules, rubric, and aggregation policy.
The resulting filter scores trajectories independently, so pairwise comparisons are required only during search.
{Starting from a blank filter, \metacurate autonomously discovers a curation strategy that achieves competitive downstream performance without task-specific handwritten criteria.}
We validate downstream utility through controlled SFT experiments on our uncurated trajectory pool and the \datasetNemotron SWE dataset, with random selection and hand-authored filtering as baselines.
{In our experiments, \metacurate achieves the highest mean \benchmarkSWEVerified score among the compared methods, including hand-authored rule-based filtering and a training-loop baseline whose search costs $3.5\times$ as much.}
\end{abstract}

\section{Introduction}
\label{sec:introduction}

{Large language models (LLMs) have made rapid progress in coding, surpassing top human experts on some competitive programming benchmarks~\citep{li2022alphacode}.}
While early work focused mainly on generating code from a task description in a single response~\citep{chen2021codex}, coding agents now solve tasks through repeated interaction with an environment~\citep{sera2026,ludwig2026swehero}.
Approaches such as ReAct~\citep{yao2023react} interleave reasoning with action, providing a framework for the repository exploration, code editing, and test-driven revision involved in agentic software engineering~\citep{jimenez2024swebench,yang2024sweagent}.
Recent work has extended this iterative problem-solving capability to the model-training process itself, so that agents construct training data, launch training runs, and use evaluation results to design the next experiment, taking on research tasks previously carried out by humans~\citep{posttrainbench2026,yu2026autotrainess,yano2025lamdagent,karpathy2026autoresearch}.
\begin{figure}[t]
\centering
\includegraphics[width=\linewidth]{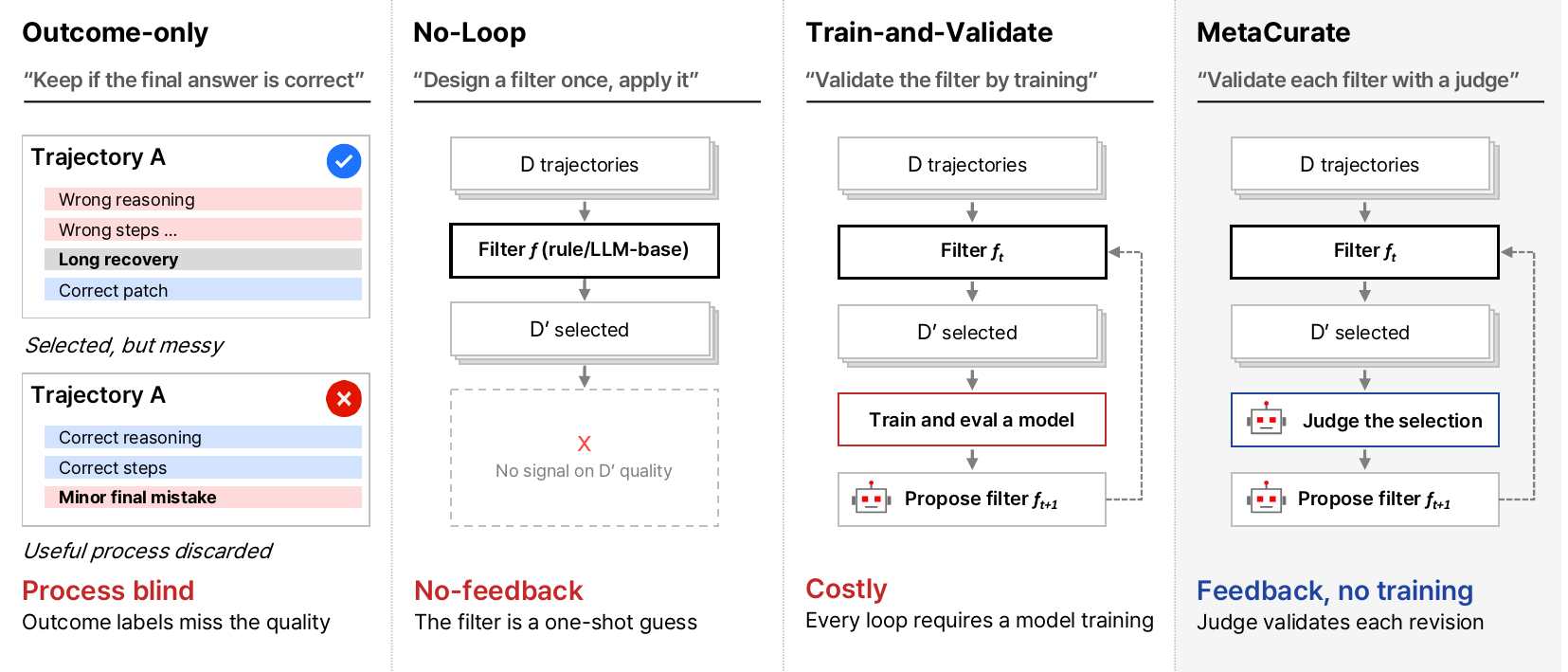}
\caption{Comparison of four approaches to selecting agent trajectories for SFT.
Outcome-only filtering can retain poor reasoning that leads to a correct patch and discard useful reasoning that ends in failure.
No-Loop curation applies a fixed rule-based or LLM-based filter without feedback on the selected data.
Train-and-Validate uses downstream training and evaluation to revise the selection, requiring a training run for each candidate.
\metacurate iteratively revises the curation filter using pairwise judge feedback, without model training during search.}
\label{fig:overview}
\end{figure}

Improving a model for agentic coding involves two separate steps: obtaining trajectories and choosing which of them to learn from~\citep{zeng2026goodagenticdata,liu2026selfplaygain}.
Supervised fine-tuning (SFT) provides a practical route to transferring capable agent behavior without repeated on-policy environment interactions during optimization~\citep{sera2026,pan2024swegym,peng2026orchard}.
The most common selection rule keeps only the trajectories whose final patch passes the tests (i.e., rejection sampling; see Fig.~\ref{fig:overview}, Outcome-only)~\citep{yang2025swesmith,nvidia2026nemotron3}.
Yet a trajectory that resolves a repository-level issue may still contain redundant exploration, malformed tool calls, or unproductive edit-and-test loops~\citep{liu2025issuefailures,zhao2026failureprocess}, and a failed one may still contain a sound diagnosis or a useful test~\citep{lv2026failforge,slinko2026srft}.
{Deciding what to keep as training data requires judging the quality of the problem-solving process together with its outcome~\citep{han2026trajcuration,raoof2026openthoughtsagent}.}

{In practice, agentic SFT curation supplements outcome labels with deterministic rules and LLM-based filtering~\citep{nvidia2026nemotron3,neohorse2026report,nanbeige2026report}.
Hand-authored rule-based filters (i.e., code-based filters) can flag recognizable failure patterns, but they are limited to behaviors specified in advance~\citep{nvidia2026nemotron3}.
LLM-based filters can assess qualitative aspects of a trajectory, but the quality of their judgments depends on the prompt and rubric (see Fig.~\ref{fig:overview}, No-Loop)~\citep{pang2024ds2,kwok2026llmverifier}.
These choices need to be validated on the target pool, yet filters are often written once and applied without such a check.}

{Recent work improves the curation strategy itself automatically~\citep{curateevo2026,khan2024dataenvgym,huang2026autodp,yang2026coevolve}.
A common approach in this line of work trains a model on the selected data and uses its failures or scores to guide the next revision (see Fig.~\ref{fig:overview}, Train-and-Validate)~\citep{curateevo2026,meng2026rsibenchdata}.
These downstream results provide direct feedback on the selected data's utility, but obtaining them for each candidate requires another training run~\citep{lim2026posttrainloop,kang2026curationbench}.
This raises a complementary question: can we iteratively improve a \instr from diagnostics and comparative feedback on real trajectories without training a model inside the search loop?}

{We propose \metacurate, a training-free framework that iteratively improves an SFT data-curation \instr (see Fig.~\ref{fig:overview}, MetaCurate).
\metacurate evaluates candidate \instrs by comparing their rankings of trajectory pairs with an LLM judge's preferences for SFT usefulness.
Each pair gives the judge a reference for relative quality, which can be more discriminating than rating either trajectory on an absolute scale~\citep{zhang2026arenarl,yang2026tournament}.
Because judging pairs is much cheaper than training a model for each candidate~\citep{curateevo2026,kang2026curationbench}, the search can gather many comparisons per iteration.
Judgments on exploration pairs also guide revision: each disagreement identifies a pair the candidate orders differently from the judge and comes with a rationale.
To make selection scalable to larger datasets, we use pairwise judgments to evolve a pointwise \instr.
}

Using this feedback, \metacurate runs an agentic search over the \instr itself.
At each iteration, a proposer (i.e., LLM coding agent with access to the run directory) reads the current \instr, its misranked trajectory pairs and judge rationales, and diagnostics of criterion health, and proposes revised candidates.
A solver executes each candidate on sampled trajectories, while pairwise judgments accumulated during exploration provide feedback for subsequent revisions.
{A revision is adopted only when its ranking improves agreement with the judge on fixed selection pairs by more than expected scoring noise.}
Crucially, no target model is trained during this search: pairwise comparisons supervise only the optimization of the \instr.
Once search is complete, the final \instr independently scores every trajectory in the full pool, requiring neither a comparison partner nor an LLM pairwise judgment at deployment time.

We evaluate whether the improvements in concordance achieved by this search translate into better training-data selection.
{To this end, we generate and release a corpus of $4{,}286$ coding-agent trajectories with \modelTeacher~\citep{deepseekv4pro} from $1{,}415$ distinct GitHub issue IDs~\citep{rebench,rebenchv2,pan2024swegym}.}
On this corpus and the $17{,}220$-trajectory \datasetNemotron SWE-SFT pool~\citep{nvidia2026nemotron3}, we conduct controlled SFT experiments with random selection as a control.
{On both pools, data selected by the evolved \instr yields the best observed mean \benchmarkSWEVerified scores among the compared methods: 64.0\% and 56.5\%, respectively.}
{These results support improving the scoring instruction itself as a way to select effective training data without placing model training inside the search loop.}

\section{Related Work}
\label{sec:relatedwork}
\paragraph{Coding agents.}
A coding agent resolves repository issues by inspecting code, running shell commands, editing files, and revising its changes based on test results~\citep{yang2024sweagent}.
In a typical SWE-bench setup, the agent receives an issue description and works in a Docker container containing the repository and its dependencies; the final patch is evaluated using held-out repository tests~\citep{jimenez2024swebench}.
SFT on trajectories generated by stronger models in synthetic environments has helped open coding agents approach frontier performance~\citep{pan2024swegym,yang2025swesmith}.
However, these datasets often include multiple attempts at the same issue, and even successful trajectories can contain unnecessary steps.
Passing tests alone does not establish whether a trajectory is a useful training example~\citep{yang2026rsr,huang2026grm}.
In practice, training pipelines use \instrs to select high-quality trajectories for SFT.

\paragraph{Data curation.}
Existing curation methods use several signals to judge whether a trajectory is suitable for training.
These include test outcomes~\citep{yuan2023rft,pan2024swegym}, soft verification against reference patches~\citep{sera2026}, and handwritten transcript heuristics~\citep{nvidia2026nemotron3,kimi2025k2}.
Other methods use LLM judgments~\citep{chen2023alpagasus}.
These signals have limitations.
LLM scores can contain model-dependent errors~\citep{pang2024ds2}, rules can be redundant~\citep{li2024orthorules}, and heuristics can bias the selected data~\citep{golubev2025swerlfilters}.
At scale, many selection methods offer little improvement over random sampling~\citep{xia2024randomselect}.
Recent work therefore moves beyond fixed criteria and optimizes the curation procedure itself.
It evolves curation code using the performance of models trained on the selected data as feedback~\citep{curateevo2026}.
\metacurate jointly evolves a \instr's rules, rubric, and aggregation policy.
It evaluates candidate \instrs through agreement with a judge on {fixed selection pairs}, without training a model for each candidate.

\section{Curation Signals and Scalable Feedback}
\label{sec:diagnosis}

\subsection{Pairwise Comparison Provides a Relative Reference}
\label{sec:diagnosis-pairwise}
{Comparative feedback has been widely used in reinforcement learning to address the weak discrimination and instability of absolute scores~\citep{shao2024grpo,rafailov2023dpo,gunjal2025rubrics,huang2025rubricanchors}.}
{Comparing two trajectories directly asks which is the more useful SFT example, with each providing a reference for assessing the other.}
{This avoids requiring independently assigned quality scores to share a calibrated numerical scale~\citep{viswanathan2025rlcf}.}
{Applying comparisons throughout a large pool can nevertheless be expensive: exhaustive comparison requires $\binom{N}{2}$ judgments for $N$ trajectories.}
{Sparse comparisons and tournaments reduce this cost, but repeated comparison-based ranking still incurs evaluation expense as the pool grows~\citep{zhang2026arenarl,yang2026tournament}.}

\subsection{Pointwise Selection Scales, but Requires Validation}
\label{sec:diagnosis-pointwise}
{Evaluating trajectories individually makes pool-wide selection practical: for a fixed scoring procedure, the number of trajectory evaluations grows linearly with pool size~\citep{chen2023alpagasus,liu2024deita}.}
{Outcome labels and LLM scores offer two ways to assess individual trajectories, while training-based feedback can validate and refine the resulting selection procedure (see Fig.~\ref{fig:overview}).}

\paragraph{Outcome labels leave training value unresolved.}
{Final task success alone does not establish the training value of the full trajectory~\citep{yang2026rsr,huang2026grm}.}
{A passing run can contain redundant exploration or malformed tool calls~\citep{liu2025issuefailures,zhao2026failureprocess}, while a failed run can contain a sound diagnosis or a useful test~\citep{lv2026failforge,slinko2026srft}.}
{In our \datasetDSVPro coding-trajectory dataset (see Appendix~\ref{sec:appendix-setup}), all attempts receive the same outcome (i.e., resolved or non-resolved) on 1,275 of 1,415 issues (i.e., $90.1\%$), so the label cannot distinguish attempts within those issues.}
Among audited pairs with different outcome labels, the SFT-suitability judge prefers the failed trajectory in $28.3\%$ of comparisons.

\paragraph{LLM scores require empirical validation.}
{LLMs can assess qualitative properties beyond task success~\citep{gunjal2025rubrics,chen2023alpagasus}, but their scores alone do not establish whether a \instr selects useful training examples or produces a meaningful ranking~\citep{pang2024ds2}.}
{Defining an appropriate evaluation standard (i.e., prompt) often requires people to inspect trajectories and formulate criteria based on experience~\citep{nvidia2026nemotron3,kimi2025k2}.}
{Without explicit criteria or reference examples to anchor the scoring scale, the same trajectory may receive different scores across repeated evaluations, and different judges may disagree on which examples are preferable~\citep{kwok2026llmverifier,zheng2023judging}.}
{{Establishing reliable evaluation criteria therefore requires an iterative process of validation and refinement on real trajectories~\citep{li2024orthorules,wang2026serpo}{~(see Appendix~\ref{sec:appendix-checks})}.}}

\begin{wrapfigure}{r}{2.8in}
\vspace{-6pt}
\centering
\includegraphics{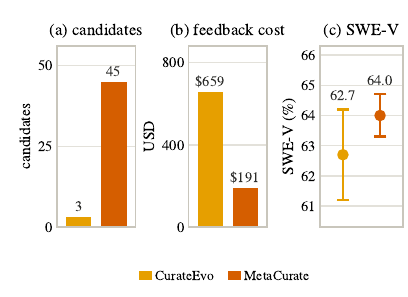}
\captionsetup{width=0.97\linewidth}
\caption{{Training feedback costs 3.5 times as much and evaluates only 3 candidates against 45.}}
\label{fig:search-budget}
\vspace{-4pt}
\end{wrapfigure}

\paragraph{Training feedback is expensive and coarse.}
\leavevmode{Training on selected data evaluates its downstream effect, but repeated validation is expensive.}
{Methods that optimize training-data selection through training--evaluation loops use downstream performance, sometimes supplemented by failure cases, to revise their curation strategies~\citep{curateevo2026,khan2024dataenvgym,meng2026rsibenchdata}.}
{Our SFT-only adaptation follows \baselineCurateEvo{}'s reported three-epoch schedule~\citep{curateevo2026}; its three candidate evaluations required 110.2 B200-hours (approximately \$659)\footnote{The compute cost is estimated using RunPod (https://console.runpod.io/)'s B200 GPU pricing.}.}
{Figure~\ref{fig:search-budget} reports the resulting run costs, not a matched-budget comparison.}
{Performance scores and failure cases help diagnose the trained model, but do not directly attribute its behavior to a particular selection rule, rubric criterion, or aggregation policy.}
{{The search therefore needs feedback that is cheaper to obtain and more directly connected to the \instr's selection decisions.}}

\takeaway{{Curation search therefore requires a selection procedure that scales to the full pool and feedback that can evaluate that procedure and suggest revisions without repeatedly training a model (Section~\ref{sec:method}).}}

\section{{\metacurate: Evolving the \InstrT}}
\label{sec:method}

\begin{figure}[t]
\centering
\includegraphics[width=\linewidth]{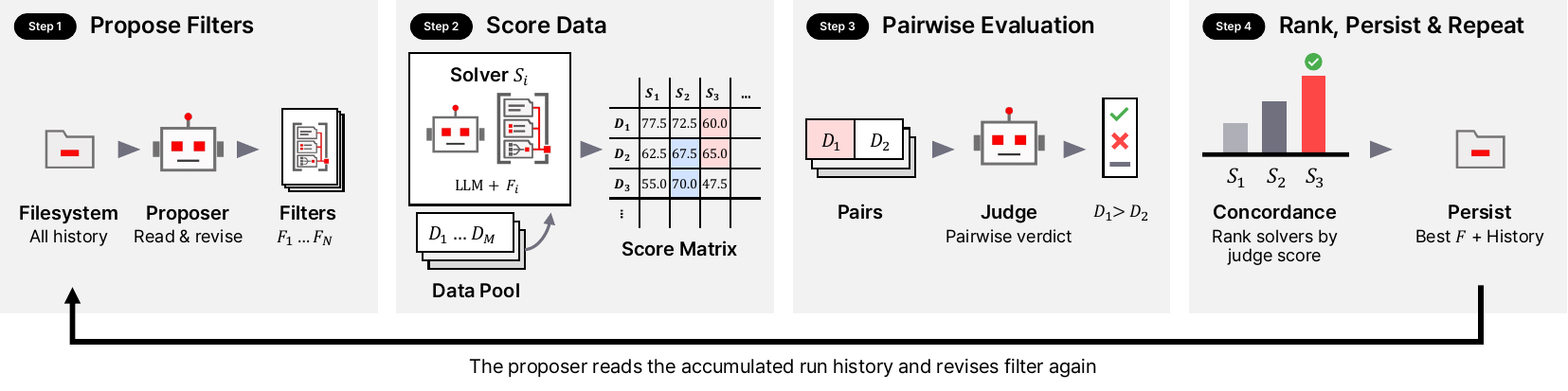}
\caption{One iteration of \metacurate.
The proposer edits the current \instr and submits candidate \instrs (i.e., rules, rubric, and policy).
The solver applies each candidate to the pool and gives every trajectory a score, which orders the pool.
The candidate whose ordering agrees most with these verdicts is kept (see Section~\ref{sec:method-promotion} for the promotion rule), and all scores, verdicts, and rationales are written to the file system for the next proposer.}
\label{fig:loop}
\end{figure}

Improving a \instr without repeated model training requires a feedback signal that can both evaluate its selections and guide its revision.
\metacurate uses pairwise judge preferences as this supervision, transferring comparison feedback into a reusable pointwise \instr.
A \instr is an executable scorer comprising three files:{\texttt{rules.py}}, \texttt{rubric.md}, and \texttt{policy.py}, which specify deterministic rules, LLM-evaluated criteria, and score aggregation.
{Appendix~\ref{sec:appendix-filter-examples} documents both seed filters, including their rubric scales, rules, and scoring policies.}
The loop evaluates candidates through comparison, edits them using actionable failure feedback, screens uncertain improvements, and applies the final \instr to the full data pool (see Fig.~\ref{fig:loop} and Algorithm~\ref{alg:metacurate}).

\subsection{{Pairwise Judgments for \InstrT Evaluation}}
\label{sec:method-objective}

We evaluate each candidate \instr\ by comparing its trajectory rankings with the judge's pairwise preferences for SFT usefulness, rather than by training a model on the data it selects~\citep{curateevo2026,khan2024dataenvgym}.
This agreement serves as a proxy for curation quality during search, while downstream SFT evaluates the utility of the selected data.

{Each \instr $F$ scores a trajectory $x$ individually, producing $s_F(x)$.
For a judged pair $(x \succ y)$, let $\kappa_F(x,y)$ equal $1$ if $s_F(x)>s_F(y)$, $\tfrac12$ for a tie, and $0$ otherwise.
Its concordance with a preference set $P$ is}
\begin{equation}
\label{eq:concordance}
{C_F(P)=\frac{1}{|P|}\sum_{(x\succ y)\in P}\kappa_F(x,y).}
\end{equation}
Our objective is to maximize $C_F(P_{\mathrm{sel}})$ subject to constraints on \instr size, criterion redundancy, label use, and permitted edits.
{The fixed selection pairs serve as an internal promotion signal, not an independent evaluation set.
Their trajectories are disjoint from the exploration subset but remain eligible for the final SFT subset.
The proposer cannot inspect the selection trajectories or pairwise verdicts; it sees aggregate promotion counts but not per-candidate selection concordance.
}
Constraints either reject a candidate before scoring or require stronger evidence for promotion (see Appendix~\ref{sec:appendix-loop}).

{Pairwise supervision is concentrated on sampled trajectories, while its reusable outcome is a pointwise scoring procedure.}
The fixed \instr then scores the full pool without further pairwise judgments.

\subsection{{\InstrT Evolution Loop}}
\label{sec:method-loop}

{To optimize the concordance objective, \metacurate repeatedly proposes revisions to the pointwise \instr and decides which candidate to promote (see Algorithm~\ref{alg:metacurate}).
Exploration pairs and their judge rationales guide proposals, while fixed selection pairs provide the promotion signal.
Keeping the selection pairs fixed provides a consistent basis for promotion, while separating them from exploration pairs avoids judging a revision on the same examples and rationales that prompted it.}

{At each iteration, the proposer uses accumulated exploration feedback and available criterion diagnostics to decide what to revise in the current \instr.
It drafts a few candidates, each changing one mechanism, and tests their behavior on a few exploration trajectories before submission.
The probe results allow the proposer to refine the drafts without requesting new judge preferences.}
{A precheck then screens the submitted candidates and permits one additional revision round before full scoring.
The solver scores valid candidates on the sampled trajectories.
The judge adjudicates exploration pairs on which candidate rankings differ, and its verdicts and rationales become feedback for later proposals.
Promotion instead compares candidates on the fixed selection pairs under the rules.
The run directory retains exploration feedback, diagnostics, and promotion records so that each iteration can build on earlier outcomes.
The retained \instr becomes the starting point for the next iteration.}

\begin{algorithm}[t]
\caption{\metacurate loop over \instrs}
\label{alg:metacurate}
\begin{algorithmic}[1]
\State \textbf{Input:} exploration set $D_{\mathrm{exp}}$,
{selection pairs $P_{\mathrm{sel}}$}, proposer $P$, iterations $N$
\State \textbf{Initialize:} current \instr $F \gets F_0$
\State \textbf{Initialize:} run directory $\mathcal{D} \gets \emptyset$

\For{$t = 1 \ldots N$}
    \State Proposer $P$ queries $\mathcal{D}$
        \Comment{prior \instrs, scores, feedback}

    \Repeat
        \State Proposer edits $F$ and probes the draft on $D_{\mathrm{exp}}$
        \State Store probe results in $\mathcal{D}$
    \Until{probe budget is spent}

    \State Proposer submits $k$ candidate \instrs $\{F_1,\ldots,F_k\}$
    \State Precheck candidates and revise rejected candidates once

    \For{$F_i \in \{F_1,\ldots,F_k\}$}
        \If{$F_i$ passes validation}
            \State $E_{i} \gets \Call{Evaluate}{F_i,D_{\mathrm{exp}},P_{\mathrm{sel}}}$
            \State $\mathcal{D} \gets \mathcal{D} \cup \{(F_i,E_i)\}$
        \EndIf
    \EndFor

    \State $F \gets \Call{Promote}{F,\mathcal{D}}$
        \Comment{{selection concordance, gated}}
\EndFor

\State \textbf{return} SFT subset selected by $F$
\end{algorithmic}
\end{algorithm}

\subsection{{Implementation Details}}
\label{sec:method-impl}

\paragraph{{Actionable failure feedback.}}
{The proposer receives the number of exploration pairs the current \instr misorders, examples of those pairs, and the judge's comparative rationales.
The examples and rationales show where the \instr disagrees with the judge and suggest revisions to its rules, rubric criteria, or aggregation.
Criterion score distributions flag criteria with small variation~\citep{wang2026serpo,guan2026evorubric}, whereas pairwise Spearman correlations flag potentially redundant rubric or rule criteria~\citep{li2024orthorules}.
Together, these diagnostics suggest which criteria to repair, remove, differentiate, or merge.}

\paragraph{{Revising and probing candidate edits.}}
{Before submitting candidates, the proposer uses the diagnostics to decide which parts of the current \instr to change.
It may repair or remove a rule that rarely fires or fires on nearly every exploration trajectory, since either pattern offers little discrimination.
Likewise, it may revise or remove a rubric criterion whose scores barely vary, or differentiate or merge criteria that appear to measure the same property.
The proposer then runs draft rules on exploration trajectories to inspect their firing rates and scores draft rubric criteria on a few exploration trajectories using the run's solver model and prompt.
The proposer can revise and probe these drafts again within its per-turn budget.
Both probing tools reject trajectories outside $D_{\mathrm{exp}}$, keeping the selection set inaccessible.}

\paragraph{{Order-consistent judgments.}}
{We present every judged pair in both AB and BA order to account for position bias in LLM judges~\citep{wang2024notfair,zheng2023judging,shi2025judgingjudges}.
We retain a preference only when both presentations identify the same winner.}

\paragraph{{Noise-aware promotion.}}
\label{sec:method-promotion}
{LLM scores can vary across calls, so we screen apparent gains using discordant-pair counts and confirm the selected candidate with fresh scoring draws~\citep{ouyang2025reasoningbank,pang2024ds2}.
This rule is intended to reduce unstable promotions; its threshold is not interpreted as a calibrated false-promotion probability.}
Candidates introducing new criterion-health defects face a stricter threshold.
Otherwise, the current \instr is retained (see Appendix~\ref{sec:appendix-loop}, Promotion).

\paragraph{Pointwise full-pool scoring.}
\label{sec:method-contract}
After search, the final \instr independently scores every trajectory in the full pool using its deterministic rules, rubric, and aggregation policy.
We rank trajectories by these scores and select the highest-scoring subset within the training-set budget, without further pairwise judging.

\section{Experiments}
\label{sec:experiments}

\subsection{Setup}
\label{sec:experiments-setup}

The main comparison evaluates the downstream utility of \metacurate{}-selected data against alternative curation methods.
Unless otherwise specified, \methodOurs refers to the \hyperref[filter:blank-seed]{blank-seed} variant of \metacurate.
The \datasetDSVPro corpus contains $4{,}286$ trajectories generated for this study by \modelTeacher using \harnessMiniSWE.
The \datasetNemotron pool contains $17{,}220$ trajectories drawn exclusively from the \harnessMiniSWE subset of the \datasetNemotron SWE-SFT release~\citep{nvidia2026nemotron3}.
We train \modelStudent for all methods, with a shared fine-tuning recipe within each pool.
The default training-set budget is $1{,}000$ trajectories; exceptions for the \baselineFullPool baseline and the \datasetDSVPro \baselineGoldPatchRecall arm are described in Appendix~\ref{sec:appendix-setup}.

{\metacurate uses \textit{\modelSolFull (\modelSol)} as proposer and \textit{\modelLunaFull (\modelLuna)} as scorer and pairwise judge.}
The LLM-based baselines also use \textit{\modelLuna} for trajectory scoring.
Downstream evaluation uses \harnessMiniSWE on all 500 \benchmarkSWEVerified instances.
{Main-comparison scores report the mean and standard deviation over three evaluations of each trained checkpoint.}
Baseline definitions and protocol details are in Appendix~\ref{sec:appendix-setup}, and training and evaluation settings are in Table~\ref{tab:hyperparams}.

\subsection{Main Comparison}
\label{sec:experiments-results}

\begin{table}[t]
\centering
\caption{Downstream performance of \modelStudent across curation methods on two pools, with a default training budget of $1{,}000$ trajectories. Scores are \benchmarkSWEVerified resolved rates (\%). The second column lists hand-authored guidance or oracle information strictly required for curation.}
\label{tab:main-comparison}
\small
\setlength{\tabcolsep}{5pt}

\setlength{\aboverulesep}{0pt}
\newcolumntype{L}[1]{>{\raggedright\arraybackslash}p{#1}}
\newcolumntype{R}[1]{>{\raggedleft\arraybackslash}p{#1}}
\begin{tabular}{L{1.85in} L{1.45in} R{0.82in} >{\centering\arraybackslash}p{0.82in}}
\toprule
Method & Task-specific Hand-authored or Oracle Information & \datasetDSVPro & \datasetNemotron \\
\midrule
\baselineBase model & --- & 53.9$\pm$2.0 & 53.9$\pm$2.0 \\
\midrule
\baselineFullPool (volume anchor) & N/A & 63.1$\pm$1.2 & 54.0$\pm$1.6 \\
\baselineRandom & N/A & 61.7$\pm$1.6 & 52.9$\pm$1.3 \\
\baselineResolved~\citep{yuan2023rft} & Execution labels & 61.9$\pm$0.7 & 54.4$\pm$0.2 \\
\baselineGoldPatchRecall~\citep{sera2026} & Reference patches & 61.7$\pm$2.4 & 53.9$\pm$0.5 \\
\baselineRuleFilter~\citep{nvidia2026nemotron3} & 7 signals, 6 thresholds & 63.1$\pm$1.4 & 54.4$\pm$0.5 \\
\baselineLLMFilter~\citep{chen2023alpagasus} & 1 prompt & 62.9$\pm$1.2 & 54.8$\pm$2.2 \\
\baselineCurateEvo~\citep{curateevo2026} & N/A & 62.7$\pm$1.5 & 53.6$\pm$0.6 \\
 \oursrow {\textbf{\methodOurs}, handwritten seed} & \hyperref[rubric:hand-written]{Handwritten 6 criteria} & 63.3$\pm$0.6 & 54.7$\pm$0.4 \\
 \oursrow {\textbf{\methodOurs}} & \hyperref[filter:blank-seed]{N/A} & \textbf{64.0$\pm$0.7} & \textbf{56.5$\pm$1.7} \\
\bottomrule
\end{tabular}
\end{table}

{\textbf{Training-free curation achieves the highest mean downstream scores.}}
Our adaptation of \baselineCurateEvo~\citep{curateevo2026} trains models inside its curation loop, while \metacurate evaluates candidate \instrs using pairwise judge preferences.
At a budget of $1{,}000$ trajectories, \modelStudent trained on data selected by the blank-seed \instr reaches $64.0\%$ on \datasetDSVPro and $56.5\%$ on \datasetNemotron, compared with $62.7\%$ and $53.6\%$ for \baselineCurateEvo, respectively (Table~\ref{tab:main-comparison}).
{\metacurate thus obtains the highest observed means among the compared methods on both pools without model training during \instr search.}
{For the handwritten-seed run, we initialize the search with the exact six-criterion \instr used to diagnose criterion failures in Appendix~\ref{sec:appendix-checks}, including its rules and scoring policy.
The blank-seed run starts from one generic criterion.
Both seed filters are documented in Appendix~\ref{sec:appendix-filter-examples}.}
\baselineRandom provides a control for subset selection: its mean scores at the same $1{,}000$-trajectory budget are $2.3$ and $3.6$ points below \metacurate on \datasetDSVPro and \datasetNemotron, respectively.
On \datasetDSVPro, the selected subset also achieves performance comparable to training on all $4{,}286$ trajectories ($63.1\%$).

\subsection{Model Scale}
\label{sec:experiments-scale}

\newsavebox{\modelScaleTable}
\begin{lrbox}{\modelScaleTable}
\begin{minipage}[t]{0.42\textwidth}
\vspace{0pt}
\captionsetup{type=table}
\centering
\caption{Transfer to \modelStudentLarge using the same training subsets as for 9B.
}
\label{tab:model-scale-27b}
\footnotesize
\renewcommand{\arraystretch}{0.93}
\setlength{\tabcolsep}{8pt}
\begin{tabular}{l c}
\toprule
Method & Resolved (\%) \\
\midrule
\baselineBase & 72.3$\pm$0.6 \\
\baselineRandom & 72.7$\pm$1.2 \\
\baselineResolved & 73.3$\pm$0.6 \\
\baselineRuleFilter & 73.1$\pm$1.0 \\
\oursrow \textbf{\methodOurs} & \textbf{74.3$\pm$0.6} \\
\bottomrule
\end{tabular}

\end{minipage}
\end{lrbox}

\begin{figure}[t]
\centering
\captionsetup[subfigure]{font=scriptsize, justification=centering, skip=3pt}
\begin{subfigure}[t]{1.75in}
\centering
\caption{Handwritten vs. blank seed}
\label{fig:role-swaps-seed}
\includegraphics{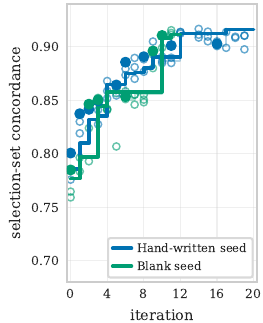}
\end{subfigure}\hfill
\begin{subfigure}[t]{1.00in}
\centering
\caption{Proposer}
\label{fig:role-swaps-proposer}
\includegraphics{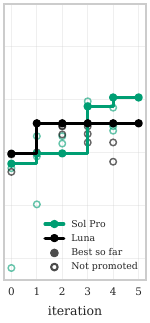}
\end{subfigure}\hfill
\begin{minipage}[t]{1.15in}
\begin{subfigure}[t]{\linewidth}
\centering
\caption{Judge}
\label{fig:role-swaps-judge}
\includegraphics{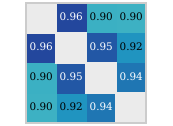}
\end{subfigure}

\smallskip
\begin{subfigure}[t]{\linewidth}
\centering
\caption{Solver}
\label{fig:role-swaps-solver}
\includegraphics{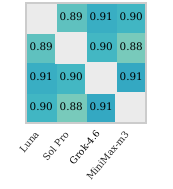}
\end{subfigure}
\end{minipage}\hfill
\begin{subfigure}[t]{1.45in}
\centering
\caption{Order flips}
\label{fig:role-swaps-order-flips}
\includegraphics{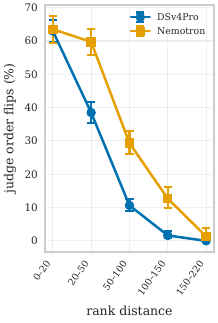}
\end{subfigure}
\caption{\textbf{Curation-loop ablations.}
(\subref{fig:role-swaps-seed})~A blank seed achieves concordance comparable to a handwritten \instr, without task-specific handwritten initialization.
(\subref{fig:role-swaps-proposer})~\modelSol achieves higher concordance than \modelLuna as proposer with the same solver and judge.
(\subref{fig:role-swaps-judge},~\subref{fig:role-swaps-solver})~Judge preferences and solver rankings largely agree across models.
Both matrices follow the model order shown in (\subref{fig:role-swaps-solver}).
(\subref{fig:role-swaps-order-flips})~Presentation-order flips are more frequent for closely ranked trajectories.}
\label{fig:role-swaps}
\end{figure}

To assess whether \metacurate's benefits extend beyond the 9B model, \modelStudentLarge is trained on the same \baselineRandom, \baselineResolved, \baselineRuleFilter, and \methodOurs subsets, without repeating \instr search.
{\methodOurs achieves a mean resolved rate of $74.3\%$ on \benchmarkSWEVerified, compared with $73.3\%$ for \baselineResolved, $73.1\%$ for \baselineRuleFilter, and $72.7\%$ for \baselineRandom (Table~\ref{tab:model-scale-27b}).}
{The gain over the compared baselines extends to 27B using the same selected data, without repeating \instr search.}

\subsection{Ablation Studies}
\label{sec:experiments-roles}

\paragraph{{Seed comparison.}}
Fig.~\ref{fig:role-swaps}(\subref{fig:role-swaps-seed}) compares the six-criterion handwritten seed analyzed in Appendix~\ref{sec:appendix-checks} with a blank seed containing one generic criterion (both shown in Appendix~\ref{sec:appendix-filter-examples}).
{For each seed, we chose the best-performing search setting before comparing the resulting filters (Appendix~\ref{sec:appendix-loop}).}
{Both starting points improve during search, reaching best observed selection concordances of $0.916$ for the handwritten seed over twenty iterations and $0.912$ for the blank seed over twelve.}
The corresponding changes in the filters' scores across search are visualized separately (see Appendix~\ref{sec:appendix-experiments}).
{The two are level on this measure, and in the main comparison the blank seed yields higher mean downstream scores than the handwritten seed} on both \datasetDSVPro ($64.0\%$ vs.\ $63.3\%$) and \datasetNemotron ($56.5\%$ vs.\ $54.7\%$; Table~\ref{tab:main-comparison}).
{These results further support obtaining useful training data without a task-specific handwritten initial \instr.}

\begin{table}[t]
\usebox{\modelScaleTable}\hfill
\begin{minipage}[t]{0.56\textwidth}
\vspace{0pt}
\centering
\footnotesize
\setlength{\tabcolsep}{3pt}
\caption{Iteration series on \datasetDSVPro.
\benchmarkSWEVerified scores are mean $\pm$ SD over three evaluations.}
\vspace{10pt}
\begin{tabular}{l cccc}
\toprule
Iteration & 3 & 6 & 9 & Final \\
\midrule
Selection $C$ & {0.845} & {0.851} & {0.856} & \textbf{0.912} \\
{Test $C$} & {0.830} & {0.839} & {0.844} & {\textbf{0.896}} \\
Resolved (\%) & 62.4$\pm$1.2 & 62.0$\pm$0.8 & 63.3$\pm$0.5 & \textbf{64.0$\pm$0.7} \\
\bottomrule
\end{tabular}
\label{tab:iteration-series}
\end{minipage}
\end{table}

\paragraph{Iteration series.}
The iteration series tests whether higher selection concordance is accompanied by better downstream performance.
\modelStudent is trained separately on the $1{,}000$ \datasetDSVPro trajectories selected at iterations 3, 6, 9, and the end of a single blank-seed run (Table~\ref{tab:iteration-series}).
As selection concordance (Selection $C$) rises from $0.845$ to $0.912$, the mean resolved rate increases overall from $62.4\%$ to $64.0\%$.
{We also score the same four filters on a separate set of 220 trajectories unused in filter search (i.e., heldout test concordance).
Test concordance (Test $C$) rises from $0.830$ to $0.896$ (Table~\ref{tab:iteration-series}).}
{Selection and test concordance improve by nearly the same amount ($+0.067$ and $+0.066$, respectively), providing evidence that the evolved \instr generalizes beyond the selection set, with no substantial selection-set overfitting apparent in this test.}
{The final \instr achieves the highest values on both measures.}
{This trend provides preliminary support for selection concordance as a search signal for downstream data utility.}
SWE-bench Verified is reserved for downstream evaluation: no benchmark instance enters \instr search, and no benchmark score is available to either the proposer or the promotion rule.
{The downstream evaluations across search iterations (Table~\ref{tab:iteration-series}) provide a separate check on whether gains in selection concordance track SFT utility.}

\paragraph{Loop component ablation.}
\textit{\modelLuna} and \textit{\modelSol} each optimize a \instr from the same blank seed, with \textit{\modelLuna} fixed as both solver and judge.
After five iterations, \textit{\modelSol} achieves higher selection concordance with the fixed judge's preferences than \textit{\modelLuna} (Fig.~\ref{fig:role-swaps}(\subref{fig:role-swaps-proposer})).
{This supports using \textit{\modelSol} as proposer to better optimize the \instr for the same solver.}
Judge agreement is measured on 100 trajectory pairs, each evaluated in both presentation orders by \textit{\modelLuna}, \textit{\modelSol}, \textit{\modelGrok}, and \textit{\modelMiniMax}.
Their pairwise agreement ranges from $0.90$ to $0.96$ {on pairs for which both judges select a consistent winner across presentation orders} (Fig.~\ref{fig:role-swaps}(\subref{fig:role-swaps-judge})).
{This broad agreement motivates using \textit{\modelLuna}, whose calls cost approximately \$0.004 each compared with \$0.14 for \textit{\modelSol}.}
Using the same final \instr optimized from the blank seed, the four solver models score 220 trajectories.
Their resulting rankings agree on $0.88$ to $0.91$ of pairs with no tied scores from either model (Fig.~\ref{fig:role-swaps}(\subref{fig:role-swaps-solver})).
{These broadly similar rankings motivate using \textit{\modelLuna}, which costs approximately \$0.004 per scoring call compared with \$0.09 for \textit{\modelSol}.}

{These results support using a stronger proposer for \instr optimization, while relying on a lower-cost model for pairwise judging and scoring with the optimized \instr.}

\begin{wraptable}{r}{0.5\linewidth}
\centering
\vspace{-12pt}
\centering
\footnotesize
\caption{200-trajectory budget on \datasetDSVPro.
}
\label{tab:low-budget}
\begin{tabular}{l c}
\toprule
Method & Resolved (\%) \\
\midrule
\baselineRandom & 58.7$\pm$1.1 \\
\baselineResolved & 59.8$\pm$1.4 \\
\baselineGoldPatchRecall & 58.1$\pm$1.2 \\
\baselineRuleFilter & 59.8$\pm$1.1 \\
\baselineLLMFilter & 58.8$\pm$0.9 \\
\baselineCurateEvo & 58.5$\pm$0.6 \\
\oursrow \textbf{\methodOurs} & \textbf{60.3$\pm$2.1} \\
\bottomrule
\end{tabular}
\vspace{-15pt}
\end{wraptable}

\paragraph{Order flips against rank distance.}
When two trajectories are close in rank, swapping their presentation order often changes the judge's preference.
Preferences are more stable for trajectories farther apart (Fig.~\ref{fig:role-swaps}(\subref{fig:role-swaps-order-flips})).
{This pattern supports judging each pair in both orders and retaining only preferences that remain consistent (Section~\ref{sec:method-objective}).}

\paragraph{Training-set budget.}
To test curation with less training data, we reuse the blank-seed \instr from the main comparison to select the top $200$ \datasetDSVPro trajectories.
A \textit{\modelStudent} model trained on this subset reaches $60.3\%$, the highest observed mean, versus $58.7\%$ for \baselineRandom and $59.8\%$ for the strongest baselines (Table~\ref{tab:low-budget}).

\section{Conclusion}
\label{sec:conclusion}

We introduced \metacurate, a training-free framework for improving trajectory \instrFs through pairwise LLM feedback.
Rather than treating curation as a fixed set of human-designed rules or validating each revision through costly model training, \metacurate iteratively revises deterministic rules, rubric criteria, and score aggregation, producing an executable \instr that can score trajectories independently at scale.
{Across two coding-trajectory datasets, the \instr evolved by \metacurate from the blank-seed \instr selects subsets that yield the highest observed mean downstream scores among the compared methods, suggesting that pairwise judgments can guide pointwise data selection.}
More broadly, our results suggest that curation itself can be systematically inspected, revised, and validated prior to training, providing a practical and interpretable approach to selecting useful SFT trajectories.

\paragraph{Limitations and future work.}
Our experiments focus primarily on whether the filters evolved by \metacurate improve downstream data selection, rather than on interpreting the curation principles that emerge during search.
An important direction for future work is therefore to analyze how the executable filter changes over time, including which rules and rubric criteria are introduced, removed, or repeatedly strengthened.
Relating these changes to downstream training outcomes could provide a more quantitative account of which trajectory properties make an example useful for SFT.
Controlled analyses of individual rules or criteria could further reveal whether the behaviors preferred by the evolved filter are subsequently reflected in the trained agent.
Ultimately, such analyses may turn \metacurate from a procedure for selecting training data into a tool for extracting empirical principles that can inform how future agent trajectories are collected, generated, and curated.

\bibliography{references}
\bibliographystyle{iclr2027_conference}

\clearpage
\appendix
\numberwithin{equation}{section}
\section{Experimental Setup Details}
\label{sec:appendix-setup}

\paragraph{\datasetDSVPro dataset.}
We collected a large, uncurated pool of coding-agent trajectories for SFT, designed to support experiments with a range of curation strategies.
To do so, \modelTeacher~\citep{deepseekv4pro} with \harnessMiniSWE is used to generate coding-agent trajectories for instances from SWE-rebench, SWE-rebench-V2, and SWE-Gym (i.e., a bash-only variant of SWE-agent)~\citep{yang2024sweagent} at temperature $0.7$.
We treat these three source datasets independently and did not deduplicate overlapping GitHub issues before collection.
Specifically, SWE-Gym shares one issue each with SWE-rebench and SWE-rebench-V2, while SWE-rebench and SWE-rebench-V2 share two issues.
We aimed to collect three attempts per source instance and produced $4{,}286$ trajectories (Table~\ref{tab:dsv4pro-sources}).
Among $1{,}415$ instance IDs, $1{,}369$ have three trajectories, 23 have two, one has a single trajectory, and 22 have six.

\begin{table}[htbp]
\centering
\caption{Source composition of \datasetDSVPro dataset.}
\label{tab:dsv4pro-sources}
\small
\begin{tabular}{lrr}
\toprule
Source dataset & Instance IDs & Trajectories \\
\midrule
SWE-rebench~\citep{rebench} & 690 & 2,100 \\
SWE-rebench-V2~\citep{rebenchv2} & 531 & 1,599 \\
SWE-Gym~\citep{pan2024swegym} & 196 & 587 \\
\midrule
Combined pool & 1,415 & 4,286 \\
\bottomrule
\end{tabular}
\end{table}

A trajectory is labeled \emph{resolved} if its submitted patch passes both the fail-to-pass and pass-to-pass tests in the instance's sandbox container, which holds for $1{,}760$ trajectories ($41.1\%$ of the pool).

\paragraph{\datasetNemotron dataset.}
To evaluate the methods on a different data distribution, we also use NVIDIA's public \datasetNemotron SWE-SFT dataset~\citep{nvidia2026nemotron3}.
Only trajectories generated with \harnessMiniSWE are retained, giving a pool of $17{,}220$ trajectories.

\begin{table}[t]
\centering
\caption{Training and evaluation settings. Shared settings are listed separately from each model's fine-tuning method and learning rate.}
\label{tab:hyperparams}
\footnotesize
\setlength{\tabcolsep}{6pt}
\begin{tabular}{l l}
\toprule
\multicolumn{2}{l}{\textbf{Training (Shared)}} \\
\midrule
Epochs & 3 \\
Schedule & cosine to $1\times10^{-6}$, warmup ratio 0.1 \\
Optimizer & AdamW, $\beta_1 = 0.9$, $\beta_2 = 0.95$, $\epsilon = 10^{-8}$, weight decay 0.01 \\
Batch & 32 \\
Sequence length & 128,000 tokens, no packing, loss on assistant turns only \\
Gradient clipping & 1.0 \\
Precision & bf16 \\
\toprule
\multicolumn{2}{l}{\textbf{Training (per model)}} \\
\midrule
9B & full fine-tuning, learning rate $1\times10^{-5}$ \\
27B & LoRA $r=64$, $\alpha=128$, dropout 0, learning rate $1\times10^{-4}$ \\
 & attention and MLP projections, FSDP full shard \\
\toprule
\multicolumn{2}{l}{\textbf{Evaluation}} \\
\midrule
Benchmark & \benchmarkSWEVerified (500 instances) \\
Sampling & temperature 1.0, top-$p$ 0.95, top-$k$ 20 \\
Output cap & 16,384 tokens per turn \\
Step limit & 250 agent steps per instance \\
Wall clock & 6 hours per instance \\
Serving & vLLM, max model length 262,144 \\
Runs & three replicates per cell, averaged; exceptions are marked in place \\
\bottomrule
\end{tabular}
\end{table}

\paragraph{Training and evaluation details.}
Table~\ref{tab:hyperparams} lists the shared settings and model-specific training configurations. We report \benchmarkSWEVerified on all 500 instances~\citep{jimenez2024swebench}.
We use the \harnessMiniSWE~\citep{yang2024sweagent} harness for the evaluation.

\paragraph{Baselines.}
For a fair comparison across curation methods, we control the training-set size by limiting each curated dataset to $1{,}000$ trajectories.
The \baselineFullPool and the \datasetDSVPro \baselineGoldPatchRecall arm are the only exceptions, as described below.

\begin{list}{\labelitemi}{\setlength{\itemsep}{0.05in}\setlength{\leftmargin}{0.2in}\setlength{\topsep}{0.02in}}

\item{\textbf{\baselineFullPool.}} This baseline trains on the entire source pool without curation.

\item{\textbf{\baselineRandom.}} A random subset of $1{,}000$ trajectories is sampled from the source pool.

\item{\textbf{\baselineResolved.}} This method retains only trajectories that pass the execution-based evaluation~\citep{yuan2023rft,pan2024swegym,yang2025swesmith}.

\item{\textbf{\baselineGoldPatchRecall.}} Following SERA~\citep{sera2026}, trajectories are retained when their reference-patch line recall is at least $0.5$.
This criterion yields $629$ training rows on \datasetDSVPro, all of which are used without relaxing the threshold or adding trajectories.
On \datasetNemotron, $1{,}000$ qualifying trajectories are sampled.

\item{\textbf{\baselineRuleFilter.}} {Following Nemotron 3 Ultra~\citep{nvidia2026nemotron3}, we apply seven checks that reject trajectories with invalid submissions, disallowed Git commands, repeated edit--test or self-revert loops, or excessive exploration with little editing.
The remaining checks flag malformed tool calls, debugging artifacts in the final patch, and edits without tests.
Our deterministic implementation fixes thresholds left unspecified in the report and does not use execution labels.}

\item{\textbf{\baselineLLMFilter.}} An LLM judge (i.e., \modelLuna) directly assesses whether each trajectory is suitable as a training example~\citep{chen2023alpagasus}.

\item{\textbf{\baselineCurateEvo.}} This method evolves a curation program using validation scores from models trained within the optimization loop~\citep{curateevo2026}.
We reserve $10\%$ of the training set for validation and use \modelLuna as the scorer.
For downstream training, we sample $1{,}000$ trajectories accepted by the resulting program.
\end{list}

\section{Loop Details}
\label{sec:appendix-loop}
\label{sec:appendix-loop-config}

\paragraph{Runs and evidence split.}
{The handwritten-seed and blank-seed \datasetDSVPro runs in Table~\ref{tab:main-comparison} share the loop and selection set but use different search settings: the handwritten run uses 20 iterations, 50 exploration trajectories, a 50-pair exploration budget, and a 0.15 complexity weight, whereas the blank run uses 12 iterations, 200 trajectories, a 200-pair budget, and no complexity weight.
This gives the minimally specified blank seed more feedback per iteration while limiting the number of search rounds.}
{\modelSolFull (\modelSol) is used for the proposer (i.e., proposing the candidate filters), while \modelLunaFull (\modelLuna) is used for the solver (scoring rubric criteria) and pairwise judge.}

{For each pool's search, we randomly sampled exploration trajectories and a disjoint selection set of 220 trajectories from that pool.}
The judge considered $6{,}673$ selection pairs in both presentation orders; the $5{,}674$ pairs with the same winner in both orders form $P_{\mathrm{sel}}$.
The proposer cannot read these rows or their judgments.
Its later promotion records expose aggregate comparison counts, not the selection trajectories or pair verdicts.
Finally, SWE-bench Verified is held out from the entire curation loop and is used only as an external test set for evaluating the downstream models trained on the selected trajectories.

\paragraph{Proposal and exploration feedback.}
At each iteration, the proposer reads the current \instr and earlier exploration feedback, probes draft rules or criteria on its exploration trajectories, and submits three candidates.
A precheck rejects invalid edits and evaluates surviving candidates against already-judged exploration pairs; the proposer can revise once before full scoring.
The solver then scores the exploration and selection trajectories with each valid candidate.
On the exploration trajectories, the judge spends the run-specific pair budget.

\paragraph{Candidate screening.}
The precheck enforces one declared mechanism per candidate and a run-wide hard size cap.
The hard caps are 9 rubric criteria and 13 \texttt{rule\_} functions; \texttt{policy.py} has no hard size cap.
The deletion limit permits at most one fewer rubric criterion and at most five fewer rules than the run's initial \instr, except for components measured to have no ranking effect.
Growth beyond the per-iteration soft cap, or a newly dead or redundant rubric criterion, requires stronger evidence at promotion rather than automatic rejection.
A criterion is dead when one score covers more than $90\%$ of rows, and a criterion pair is redundant when its tie-aware Spearman correlation exceeds $0.7$ in magnitude.
These checks reject candidates that violate hard edit or size constraints while leaving soft-cap and criterion-health concerns to the promotion test below.

\pagebreak
\paragraph{Promotion.}
{The promotion screen uses binary correctness, distinct from the half-credit concordance in Eq.~\ref{eq:concordance}: only a strict ordering in the judge's preferred direction counts as correct. Pairs tied by the judge are excluded from both measures.}
For a candidate and the current \instr, let $b$ count selection pairs ordered correctly only by the candidate and $c$ count pairs ordered correctly only by the current \instr:
\begin{equation}
\label{eq:bc}
b=\#(\text{candidate only correct}),\qquad
c=\#(\text{current \instr only correct}).
\end{equation}
{A filter's own tie or missing score is not a correct ordering for this screen, even though its own tie earns half credit in concordance.}
{An ordinary candidate needs at least five discordant pairs and is screened using a one-sided binomial-tail statistic:}
\begin{equation}
\label{eq:pvalue}
{p(b,c)=\Pr[B\ge b],\qquad B\sim\mathrm{Binomial}(b+c,\tfrac12).}
\end{equation}
{The implementation evaluates this tail exactly for small discordant counts and uses a normal approximation for larger counts.}
{We start from a nominal one-sided $z=2$ threshold, adjust it across candidates within an iteration using the \v{S}id\'ak formula, and add one $z$ unit for soft-cap or new criterion-health defects.}
{Because selection pairs share trajectories and the set is reused across iterations, the nominal threshold does not provide a guaranteed error rate for the full search.}
A candidate no larger in any size component and smaller overall can instead qualify as a simplification when $b\ge c$, but an eligible improvement takes priority.
Among eligible improvements, the highest-concordance leader must also be distinguishable from competing eligible candidates; otherwise size and then edit size break the tie, or the current \instr is kept.
The selected leader is scored again with fresh LLM draws and promoted only if its new comparison favors it and the pooled comparison still clears its threshold.

\paragraph{Stopping.}
Each run stops at its configured iteration limit or after five consecutive iterations without promotion once at least four iterations have run.

\section{Checking Hand-Written Filters on Data}
\label{sec:appendix-checks}
\label{sec:appendix-triangulation}
\label{sec:appendix-loo}

\begin{figure}[t]
\centering
\includegraphics[width=\linewidth]{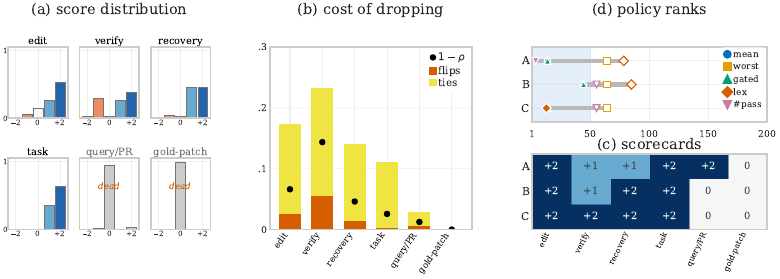}
\caption{{A plausible hand-written filter need not behave as intended.
(a)~Two of six criteria scarcely vary over 200 trajectories.
(b)~Dropping each criterion changes pairwise orderings (bars) and rank distance
$1-\rho$ (dots); the nearly constant criteria change almost nothing.
(c,d)~The same criterion scores for trajectories A--C yield different rankings
under five aggregation policies. Shading marks the top 50; filled markers mean
selection probability at least one half. In (d), gated denotes gated mean and
lex denotes verif-lex.}}
\label{fig:dead-axes}
\label{fig:dead-axes-loo}
\label{fig:policy-disagreement}
\end{figure}

{A human hand-written rule or rubric can read plausibly yet behave differently when applied to trajectories.
We therefore check whether its criteria affect rankings and which examples its final policy selects for training.}

\textbf{Dead criteria.} {In a hand-written six-criterion \instr~(\hyperref[rubric:hand-written]{{see the full rubric}}), two criteria give almost the same score to every one of 200 trajectories. Removing either barely changes their ranking (see Fig.~\ref{fig:dead-axes} (a) and (b)). Written distinctions alone do not ensure that a criterion can influence selection.}
\begin{wrapfigure}[13]{r}{1.8in}
\centering
\includegraphics[width=1.45in]{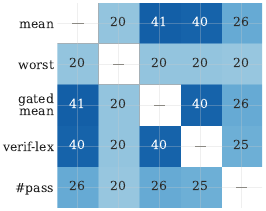}
\captionsetup{width=\linewidth,skip=3pt}
\caption{Top-50 policy overlap with shared~scores.}
\label{fig:policy-overlap}
\end{wrapfigure}

{The Gold-Patch Relation criterion is effectively inactive here because reference gold patches were not supplied to the solver and judge, leaving it without the information its rubric requires.}

\textbf{Score aggregation.} {The \texttt{policy.py} component also matters: it combines rubric scores and rule outcomes into a final score that determines which trajectories are selected for training.
With criterion scores held fixed, five plausible aggregation policies rank the same three trajectories differently (see Fig.~\ref{fig:dead-axes} (c) and (d)), and their expected top-50 selections overlap by only $19.7$ to $40.9$ of 50 trajectories (see Fig.~\ref{fig:policy-overlap}).
The policy must therefore be checked on the pool, alongside the rules and rubric that supply its inputs.}

{These cases show why even a hand-crafted \instr must be tested
on real trajectories. Its criteria may contribute little to the ranking, and its aggregation policy can change which examples are selected.
Inspecting scores, rankings, and selections by hand for every candidate is costly; we therefore automate the evaluation and revision loop.}

\section{Filter Evolution}
\label{sec:appendix-experiments}
\label{sec:experiments-halo}

\begin{figure}[p]
\centering
\includegraphics[width=\linewidth,height=0.84\textheight,keepaspectratio]{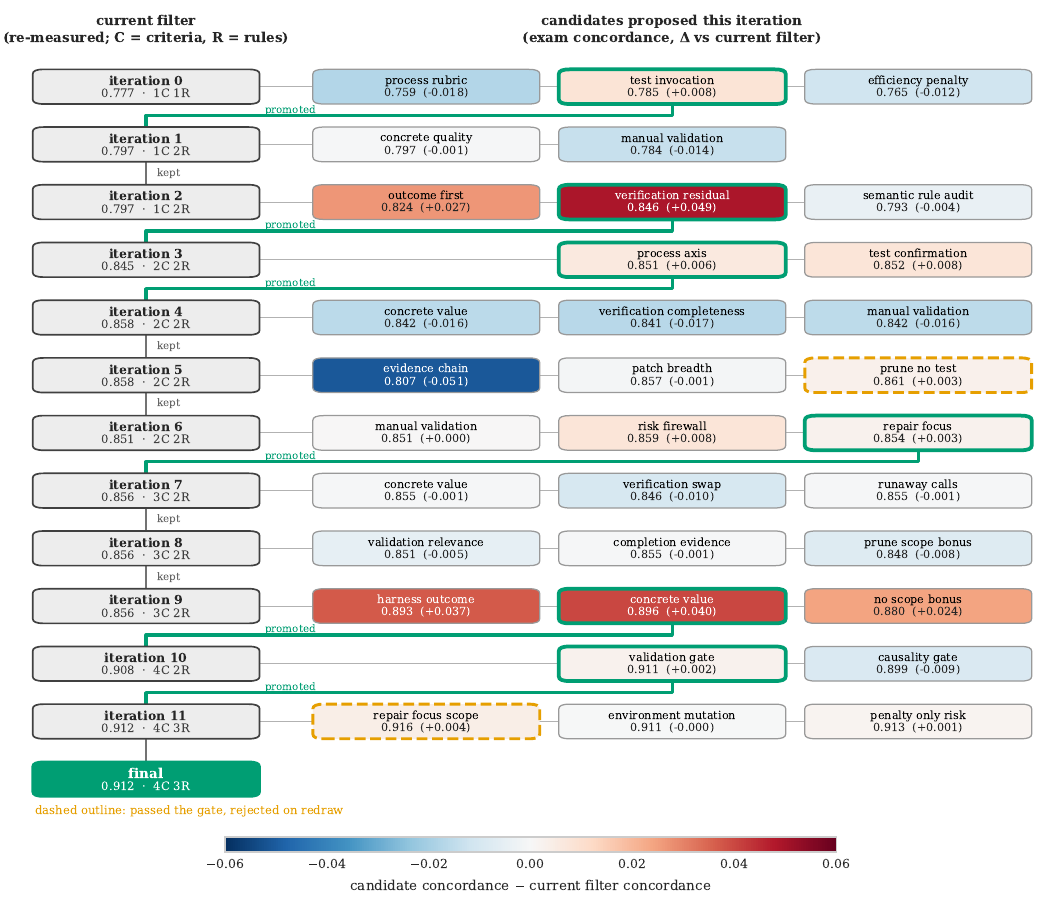}
\caption{Search history from a blank seed over twelve iterations.
Each row shows the current \instr on the left and proposed candidates on the right; C and R denote the numbers of criteria and rules.
Empty slots indicate candidates that did not pass precheck (Algorithm~\ref{alg:metacurate}).
Red and blue indicate positive and negative concordance differences from the current \instr, with darker colors indicating larger differences.
Green outlines mark promoted candidates.
Orange dashed outlines mark eligible candidates in iterations where the provisionally selected candidate failed re-evaluation, leaving the current \instr unchanged.}
\label{fig:filter-beam-ab2}
\end{figure}

\begin{figure}[p]
\centering
\includegraphics[width=\linewidth,height=0.84\textheight,keepaspectratio]{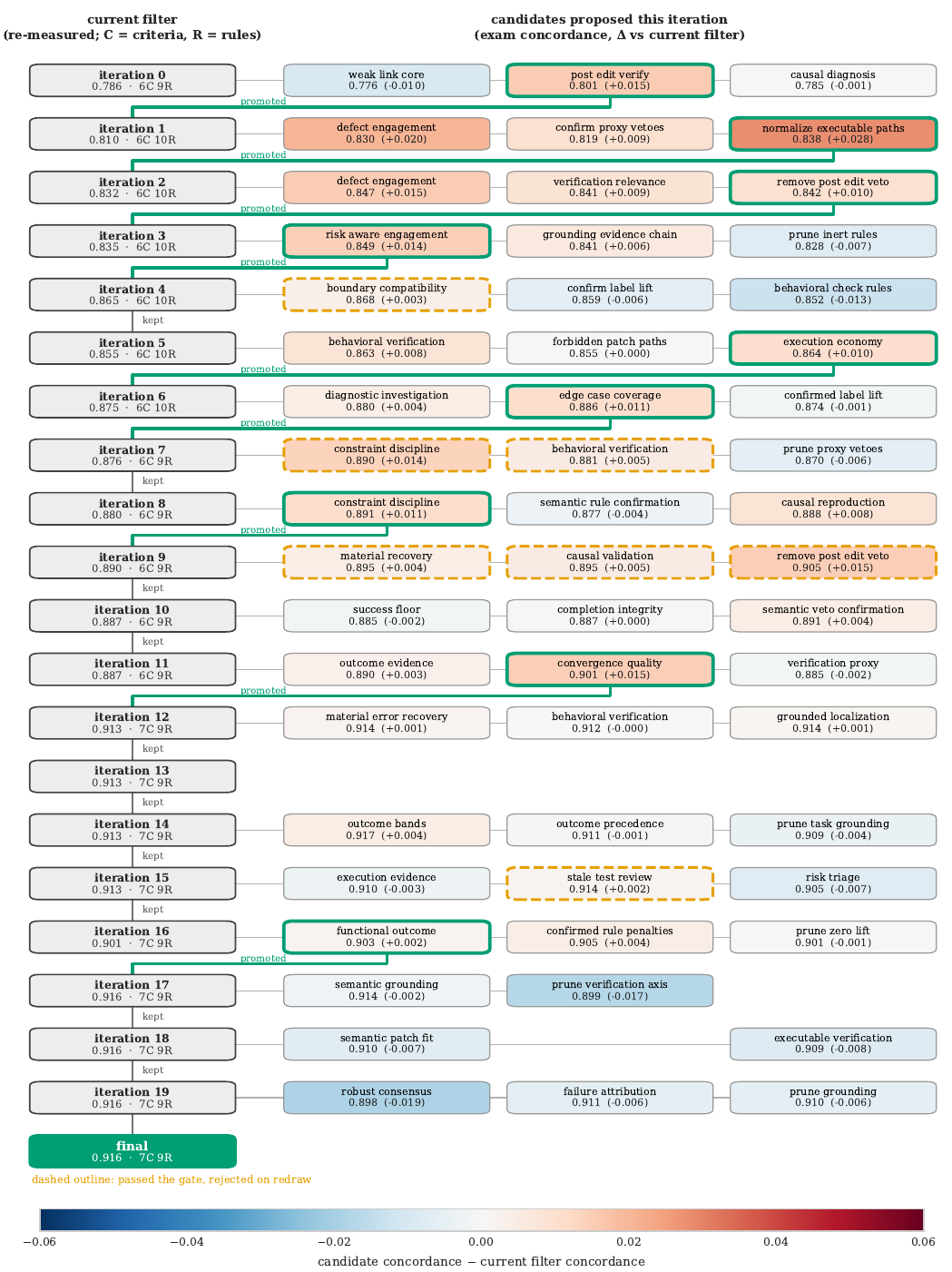}
\caption{Search history from a handwritten seed over twenty iterations.
Notation and color scale match Fig.~\ref{fig:filter-beam-ab2}.
Mostly pale cells indicate few large departures from the current \instr, consistent with greater inertia than in the blank-seed run.}
\label{fig:filter-beam-hand}
\end{figure}

Blank and handwritten seeds show different search patterns (see Fig.~\ref{fig:filter-beam-ab2} and~\ref{fig:filter-beam-hand}).
Darker red and blue indicate larger positive and negative score differences from the current \instr.
The blank-seed run produces more pronounced departures in both directions, whereas the handwritten-seed run has mostly pale cells and fewer large score differences.
This pattern is consistent with greater inertia around an established handwritten \instr, while a blank seed leaves more room for substantial changes in scoring behavior.

\clearpage
\section{Role Prompts}
\label{sec:appendix-role-prompts}

The boxes give compact excerpts of the prompts used by the three LLM roles.
Braced fields are filled at runtime; omitted evidence blocks and operational
instructions are indicated by brackets. The solver evaluates each criterion in a separate LLM call containing the rubric preamble and that criterion, reducing conflation across assessment dimensions~\citep{kwok2026llmverifier,viswanathan2025rlcf}.
Each pair is shown to the judge in both orders.
The role prompts were not extensively tuned for these experiments; nevertheless, the reported runs show effective filter evolution with this fixed configuration.

\begin{promptbox}{Proposer prompt}
\begin{lstlisting}
[System] You are an optimization engineer improving a filter that
decides whether an agent trajectory is useful training data.
The filter is a continuous scorer: decide() returns a score in [0, 100]
for every trajectory. Improve its ranking of trajectory pairs.
Selection uses fixed pairs you cannot inspect. A candidate replaces
the current filter only when it orders significantly more of those
pairs correctly than the current filter.

[User] Produce {n_candidates} candidate filter revisions. Each
candidate supplies complete replacements for rubric.md, rules.py,
and policy.py. Use the visible misordered exploration pairs, judge
rationales, and criterion diagnostics to identify ordering failures.
Return one JSON object containing a diagnosis and candidates with
names, target pairs, per-file rationales, and replacement file text.

[Runtime evidence: current filter files, misordered pairs, critic
reports, criterion diagnostics, prior diagnoses, and promotion records]
\end{lstlisting}
\end{promptbox}

\begin{promptbox}{Solver prompt}
\begin{lstlisting}
[System] You are a strict evaluator scoring an agent trajectory
against a rubric. Honor the rubric text verbatim; do not invent
criteria. Output exactly one JSON object.

Score the following trajectory on the scale +2 / +1 / 0 / -1 / -2.
Give one short line of concrete evidence for each score.

RUBRIC (verbatim): {rubric preamble and one criterion}
Task: {task}
Metadata: {permitted metadata fields}
Action histogram: {histogram}
Full transcript: {turns_block}
Final submitted patch: {submission}
{optional reference blocks, if the criterion requests them}

Return JSON with scores keyed by criterion title, each containing
"score" and "evidence", plus a one-sentence "summary".
\end{lstlisting}
\end{promptbox}

\begin{promptbox}{Pair-judge prompt}
\begin{lstlisting}
You are comparing two SWE agent trajectories as candidate SFT
training data. Which one would a strong SWE model learn more good
behavior from imitating? Consider diagnosis quality, patch quality,
verification discipline, and efficiency. Judge the process, not
just the outcome.

TRAJECTORY A ({id_a})
Task: {task_a}
Harness metadata: {meta_a}
Transcript: {turns_a}
Final submitted patch: {submission_a}

TRAJECTORY B ({id_b})
Task: {task_b}
Harness metadata: {meta_b}
Transcript: {turns_b}
Final submitted patch: {submission_b}

Return exactly: {"winner": "A" | "B" | "TIE",
                 "rationale": "one sentence"}
\end{lstlisting}
\end{promptbox}

\clearpage
\section{Seed Filter Examples}
\label{sec:appendix-filter-examples}
{The handwritten run in experiment section starts from exactly the six-criterion filter examined in Appendix~\ref{sec:appendix-checks}; the blank run starts from one generic criterion. The boxes below document both seed filters, including their rubric anchors, rules, and scoring policies.}
\begingroup
\tcbset{coltext=black,colbacktitle=promptback}
\newcommand{\rubricitem}[2]{
  \par\vspace{1.5pt}
  {\raggedright\leftskip=0.7em\hangindent=2.7em\hangafter=1
   \noindent\makebox[2.7em][l]{\textbf{#1.}}#2\par}
}
\begin{promptbox}{\textcolor{black}{Hand-written six-criterion rubric in Appendix Sec.~\ref{sec:appendix-checks}}}
\phantomsection
\label{rubric:hand-written}
\footnotesize
We are filtering agent rollouts that attempt to solve a single software-repair task (typically an issue / PR description). Each criterion is scored on a symmetric +2 / +1 / 0 / -1 / -2 scale. Anchors describe both what a \emph{good} and a \emph{bad} trajectory looks like so the same rubric supports FP and FN critics.

\vspace{6pt}
Score every criterion and cite one line of concrete evidence per non-zero score. When the evidence a criterion needs is absent or ambiguous, the score is 0 --- doubt is 0, never a guessed +1 or -1.

\vspace{6pt}
Navigation-only actions (\texttt{cd}, \texttt{pwd}, \texttt{ls}) are structural bookkeeping --- do NOT count them as "information gain" or "repetition"; look at the \emph{content} actions (\texttt{grep}, \texttt{read/cat}, \texttt{edit/apply\_patch}, \texttt{test/pytest}, \texttt{python}, \texttt{run\_tests}, \texttt{sed}, \texttt{awk}) instead.

\vspace{6pt}
Two criteria read reference material that ships with the task (\texttt{metadata.problem\_statement}, {\texttt{metadata.gold\_patch}}). They judge the \emph{task} and the agent's \emph{relation} to the reference --- they must never be scored by similarity alone, and when the reference fields are missing the score is 0 with a note saying the reference was unavailable.

\medskip\noindent\textbf{Edit Justification.} Do the applied edits address the task described in the trajectory?
\rubricitem{$+2$}{Every edit maps to a specific claim about the bug/feature; edits are surgical (few, targeted hunks).}
\rubricitem{$+1$}{Edits address the task but include some unrelated churn.}
\rubricitem{~~~$0$}{Edits partially address the task; hard to tell if the fix is on the right layer --- doubt is 0.}
\rubricitem{$-1$}{Edits are speculative --- the agent guesses and reverts multiple times.}
\rubricitem{$-2$}{Edits contradict the task, or the submitted patch is empty / trivial ($<400$ bytes for a non-trivial task) with no verification.}

\medskip\noindent\textbf{Verification Completeness.} Does the trajectory actually verify that the change works?
\rubricitem{$+2$}{Runs relevant tests (\texttt{pytest tests/...}, \texttt{python -m pytest}, project's build/lint) AFTER the final edit and observes success.}
\rubricitem{$+1$}{Runs some tests (smoke / lint) but not the full relevant suite.}
\rubricitem{~~~$0$}{Runs tests but only intermittently; verification is unclear --- doubt is 0.}
\rubricitem{$-1$}{Runs tests early but not after the final edit, OR declares completion without running any test after the edit.}
\rubricitem{$-2$}{Declares completion (\texttt{task complete}, \texttt{all done}, submits patch) with no test/build invocation anywhere in the trajectory.}

\medskip\noindent\textbf{Error Recovery.} When commands fail, does the agent adapt?
\rubricitem{$+2$}{Every non-zero exit or traceback is followed by a diagnostic action that materially changes the next step.}
\rubricitem{$+1$}{Most errors are handled; occasional retry-without-change.}
\rubricitem{~~~$0$}{Errors present but mostly benign (search misses etc.), or too few errors to judge --- doubt is 0.}
\rubricitem{$-1$}{Same failing command is repeated without a plan change.}
\rubricitem{$-2$}{Persistent unrecovered errors (repeated identical failing commands or unhandled tracebacks in the last third of the trajectory).}

\medskip\noindent\textbf{Task Grounding.} Does the agent read and internalize the task before acting?
\rubricitem{$+2$}{Agent quotes / paraphrases the task and localizes the affected symbol within the first $\sim$5 content actions.}
\rubricitem{$+1$}{Agent references the task but jumps to editing without full localization.}
\rubricitem{~~~$0$}{Task is acknowledged but not clearly used to steer actions; or the transcript is too truncated to tell --- doubt is 0.}
\rubricitem{$-1$}{Agent starts editing before demonstrating any understanding of the task.}
\rubricitem{$-2$}{Agent hallucinates a task or edits files unrelated to the request.}

\medskip\noindent\textbf{Query/PR Quality.} Is the task itself worth training on: is the problem statement specific enough to act on, {and is the reference (gold) patch a coherent fix for what the statement describes?}
\rubricitem{$+2$}{The problem statement names concrete symptoms (error text, reproducible example, expected vs actual behavior) {AND the gold patch, when present, plainly addresses that described behavior.}}
\rubricitem{$+1$}{The statement is actionable but underspecified in places (e.g. no repro, vague expected behavior); {the gold patch still matches its topic.}}
\rubricitem{~~~$0$}{Quality cannot be judged --- the statement {or gold patch} is missing, truncated, or generic boilerplate. Doubt is 0.}
\rubricitem{$-1$}{The statement is ambiguous enough that two reasonable fixes would differ (no repro, no expected behavior, mixed concerns), {or the gold patch fixes something the statement never mentions.}}
\rubricitem{$-2$}{The statement is incoherent, self-contradictory, {or the gold patch is unrelated to it (e.g. pure refactor / version bump against a bug report)}: any agent behavior on this task is unlearnable noise.}

\medskip\noindent{\textbf{Gold-Patch Relation.} Does the agent's submitted diff engage the same defect the reference patch fixes --- same file/function region, or a defensible alternative route?}
\rubricitem{{$+2$}}{{The submitted diff edits the same file(s) and the same function/region as \texttt{metadata.gold\_patch}, or the transcript shows the agent found the gold patch's site and chose a different layer with stated reasons that the tests then support.}}
\rubricitem{{$+1$}}{{Same file(s) as the gold patch but a different region, and the edit plausibly reaches the same behavior; or a different file with an explicit, test-backed argument connecting it to the defect.}}
\rubricitem{{~~~$0$}}{{\texttt{metadata.gold\_patch} is absent, or either diff is unreadable --- the relation cannot be judged. Doubt is 0. Never score similarity for its own sake: a textually different but behaviorally equivalent fix is not a defect.}}
\rubricitem{{$-1$}}{{The submitted diff shares no files with the gold patch and the transcript offers no evidence (test run, traced call path) connecting the edited site to the defect.}}
\rubricitem{{$-2$}}{{The submitted diff works around the defect instead of engaging it --- hardcodes expected output, edits/disables the failing test, or patches a caller to mask the symptom the gold patch fixes at the~source.}}

\end{promptbox}
\endgroup

\begingroup
\tcbset{coltext=black,colbacktitle=promptback}
\newcommand{\filteranchor}[2]{
  \par\vspace{1.5pt}
  {\raggedright\leftskip=0.7em\hangindent=2.7em\hangafter=1
   \noindent\makebox[2.7em][l]{\textbf{#1.}}#2\par}
}
\begin{promptbox}{\textcolor{black}{Handwritten seed: rules and policy}}
\phantomsection\label{filter:hand-seed}
\footnotesize
The rubric is printed in the preceding box. The handwritten run starts from that exact six-criterion filter, including the same \texttt{rules.py} and \texttt{policy.py}.

\medskip\noindent\textbf{Rules.} Nine deterministic rules inspect the transcript, submitted diff, or grading labels:
\begin{itemize}\setlength{\itemsep}{1pt}\setlength{\parsep}{0pt}
\item A missing or invalid git-diff submission, disallowed git command, eight returns to editing after tests or three self-reverts, or exploration with no edit is rejected. Exploration also fires after at least 20 actions when at least 85\% are reads or searches and at most two are edits.
\item A malformed-command rate above 10\%, debug artifacts in added patch lines, or an edit with no test invocation also fires a rejection rule.
\end{itemize}
\noindent\textbf{Policy.} Let $m$ be the mean of the available rubric scores, $r$ the number of firing rejection rules, $\ell$ the positive-label indicator, and $w$ indicate any rubric score of $-2$. The score is $\operatorname{clip}_{[0,100]}(30+15m+25\ell-15\min(r,2)-10w)$; a score of at least 65 receives \texttt{ACCEPT}. Ranking uses the continuous score.
\end{promptbox}

\begin{promptbox}{\textcolor{black}{Blank seed: rubric}}
\phantomsection\label{filter:blank-seed}
\footnotesize
The blank run starts from a single generic criterion. A nonzero score requires one line of evidence; absent or ambiguous evidence receives $0$.

\vspace{6pt}
\medskip\noindent\textbf{Overall Training Value.} How valuable is this trajectory as supervised fine-tuning data?
\filteranchor{$+2$}{clearly among the most valuable rollouts available.}
\filteranchor{$+1$}{somewhat more valuable than a typical rollout.}
\filteranchor{~~~$0$}{typical, or the transcript does not support a judgement --- doubt is 0.}
\filteranchor{$-1$}{somewhat less valuable than a typical rollout.}
\filteranchor{$-2$}{clearly among the least valuable rollouts available.}
\end{promptbox}

\begin{promptbox}{\textcolor{black}{Blank seed: rules and policy}}
\phantomsection\label{filter:blank-seed-policy}
\footnotesize
\noindent\textbf{Rule.} The only rule rejects a trajectory with zero turns. It contains no task-specific repair heuristic.

\vspace{6pt}
\noindent\textbf{Policy.} With criterion score $s$, the raw score is $50+12.5s$ plus a rule adjustment capped to $[-15,15]$; the final score is clipped to $[0,100]$, and \texttt{ACCEPT} requires at least 50. The policy lists a $-15$ weight for \texttt{no\_turns}, but the solver passes the function name \texttt{rule\_no\_turns}; a fired rule therefore takes the default \texttt{REJECT} adjustment of $-8$.
\end{promptbox}

\endgroup

\clearpage

\end{document}